\documentclass[a4paper, 10pt, conference]{ieeeconf}      

\IEEEoverridecommandlockouts                              
\usepackage{graphicx} 
\usepackage{float}

\title{
DeepScores and Deep Watershed Detection: current state and open issues}

\author{
Ismail Elezi *\\
ZHAW Datalab \\
Winterthur, Switzerland \\
{\tt\small elez@zhaw.ch}
\and
Lukas Tuggener *\\ \thanks{*Equal contribution}
ZHAW Datalab \\
Winterthur, Switzerland \\
{\tt\small tugg@zhaw.ch}
\and
Marcello Pelillo\\
Ca' Foscari University\\
Venice, Italy \\
{\tt\small pelillo@unive.it}
\and
Thilo Stadelmann\\
ZHAW Datalab\\
Winterthur, Switzerland \\
{\tt\small stdm@zhaw.ch}
}

\begin{document}

\maketitle
\thispagestyle{empty}
\pagestyle{empty}

\begin{abstract}

This paper gives an overview of our current Optical Music Recognition (OMR) research. We recently released the OMR dataset \emph{DeepScores}  as well as the object detection method \emph{Deep Watershed Detector}. We are currently taking some additional steps to improve both of them. Here we summarize current and future efforts, aimed at improving usefulness on real-world task and tackling extreme class imbalance. 

\end{abstract}

\section{INTRODUCTION}
The accurate localization and classification of musical symbols is a key component in every functioning Optical Music Recognition (OMR) system \cite{Rebelo}. In pursuit of our goal of advancing the state of the art in optical music symbol detection we have created the large \emph{DeepScores} \cite{DeepScores} dataset of synthetic music scores together with ground truth to enable the training of very deep neural networks. Additionally, we created a custom object detection method called Deep Watershed Detection \cite{dwd}, that is designed to work particularly well on optical music notation data. Both these contributions currently carry some drawbacks and flaws that hamper performance and usability. In this paper, we give an overview of our current as well as planned efforts to alleviate these issues.

\section{UPDATES TO THE \emph{DeepScores} dataset}

\subsection{Shortcomings of the initial release}
At its initial release, \emph{DeepScores} had two main weaknesses: first, it was fully geared towards our application in conjunction with Audiveris; many common symbols that where not interesting in that context have been omitted, which severely limited the usability of \emph{DeepScores} in other contexts. Second, \emph{DeepScores} consist only of synthetically rendered music sheets, since labelling hundreds of thousands of music sheets by hand is prohibitively expensive. However, the common use case for OMR is scans or even photos of music sheets. This discrepancy can lead to severe performance drops between model training and actual use.

\subsection{Enhanced character set}
In an effort to make \emph{DeepScores} more universally usable we created a new version---called \emph{DeepScores-extended}---containing annotations for a far greater number of symbols. According to our knowledge and discussions with other members of the community, no crucial symbols are missing from the \emph{DeepScores-extended} annotations. The full list of supported symbols is available online\footnote{$tuggeluk.github.io/deepscores\_syms\_list$}.

\subsection{Richer musical information}
\begin{figure}[ht]
	\begin{center}
    \includegraphics[width=0.5\textwidth]{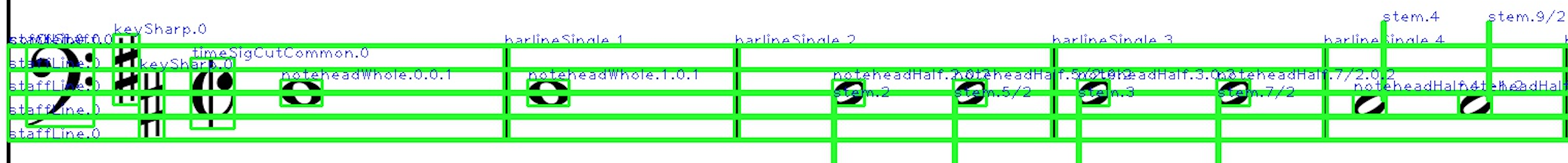}
    \caption{Small piece of music notation with DeepScores-extended annotations overlayed. The naming is either classname.onset or classname.onset.relativecoordinate.duration, depending on availability.}
    \label{fig:annotations_extended}
    \end{center}
    \vspace{-0.8cm}
\end{figure}

While the interest of the authors lies in the detection of musical symbols, this task is not the full problem of OMR. The reconstruction of semantically valid music from detected symbols is at least as challenging as the detection. To enable research focused on reconstructing higher-level information, we have added additional information to the \emph{DeepScores} annotations. Every labeled object now has an \emph{onset} tag that tells the start beat of the the given object. All noteheads additionally have their relative position on the staff as well as their duration in their annotation (see Figure \ref{fig:annotations_extended}).

\subsection{Planned improvements}
\begin{figure}[ht]
	\begin{center}
    \includegraphics[width=0.35\textwidth]{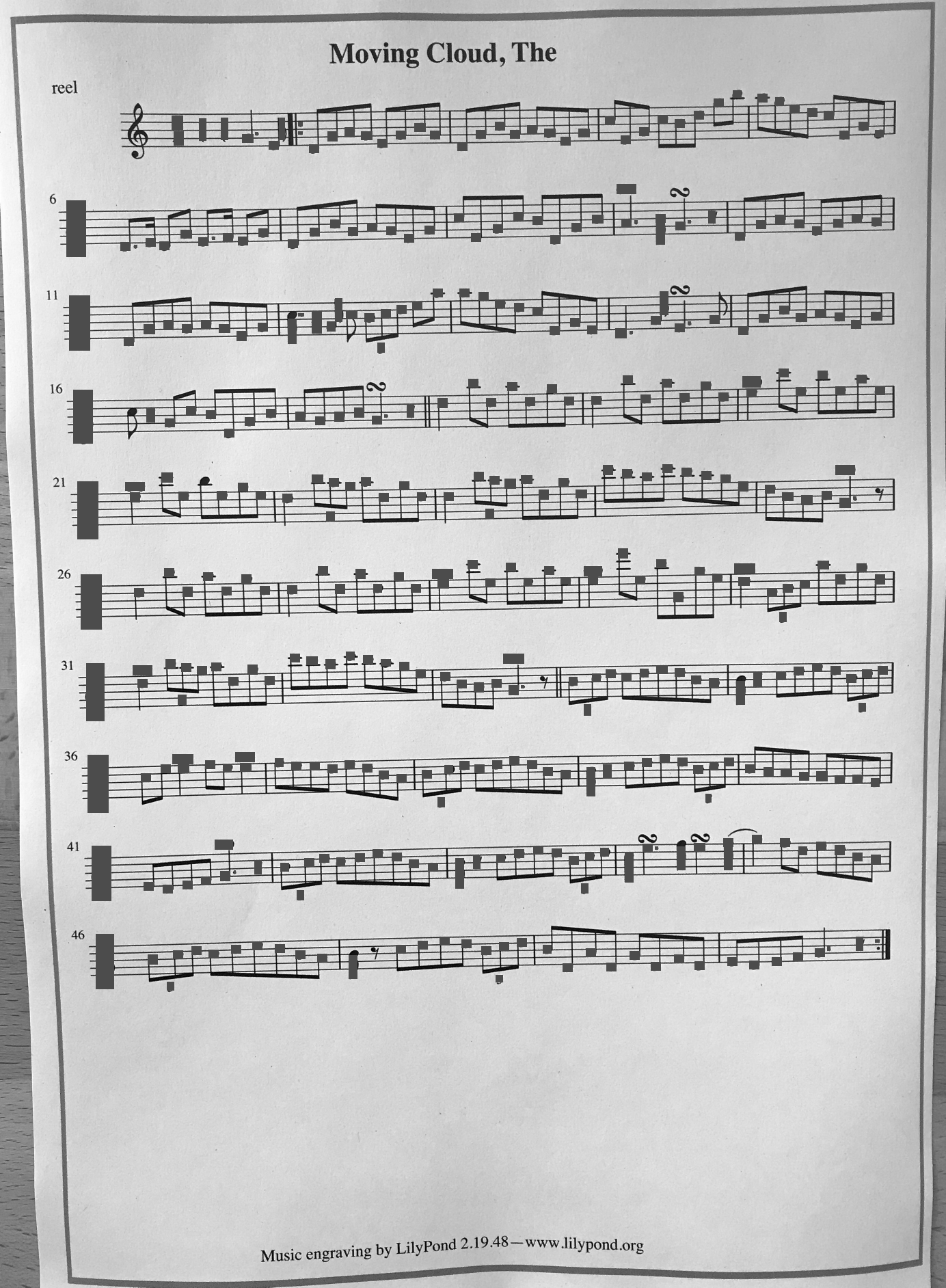}
    \caption{Preliminary results of our model (grey boxes) on a photo of a printed sheet. While not perfect (for example, our model misses the clef in the first row), they already look promising.}
    \label{fig:data_scanned}
    \end{center}
    \vspace{-0.8cm}
\end{figure}

A drawback of the \emph{DeepScores} dataset is that it is synthetic. We are currently working on a much smaller dataset, meant for transfer-learning, that consists of pages originally taken from \emph{DeepScores} that are printed and then digitized again. Then, through a global centering and orientation alignment of the scan, the original annotations are made valid again for the scanned version. We use different printers, scanners, cell-phone cameras, and paper qualities to make the noise introduced by this process resemble the real world use case as much as possible. Naively training a Deep Watershed Detector on this new dataset, we observed that the detector was unable to find anything on the testing set despite that the loss function converged. This led us to believe that severe overfitting is going on, and we were able to get promising results by simply adding l2-regularization and performing more careful training (see Figure \ref{fig:data_scanned} for a qualitative result of the detector on the new dataset).

\section{FURTHER RESEARCH ON DEEP WATERSHED DETECTION}
\subsection{Augmenting inputs}

\emph{DeepScores}, unlike many academic datasets, is extremely unbalanced. In fact, the most common class (notehead black) contains more symbols than the rest of the classes combined, while the top $10$ classes contain more than $85$\% of the symbols. However, some of the rare symbols are important and simply dismissing them might lead to semantic problems during the reconstruction of valid music in some digital format. Initially, we tried to solve the problem by using a weighted loss function which penalizes more severely the mistakes on the rare symbols, but to no avail. In \cite{dwd} we conjecture that the inbalance is so extreme that simply weighting the loss function leads to numerical instability, while at the same time the signal from these rare symbols is so sparse that it will get lost in the noise of stochastic gradient descent during the training: many symbols will be present only in a tiny fraction of mini batches. Both of these problems do not get solved by a weighted loss function. 

\begin{figure}[t]
	\begin{center}
    \includegraphics[width=0.42\textwidth]{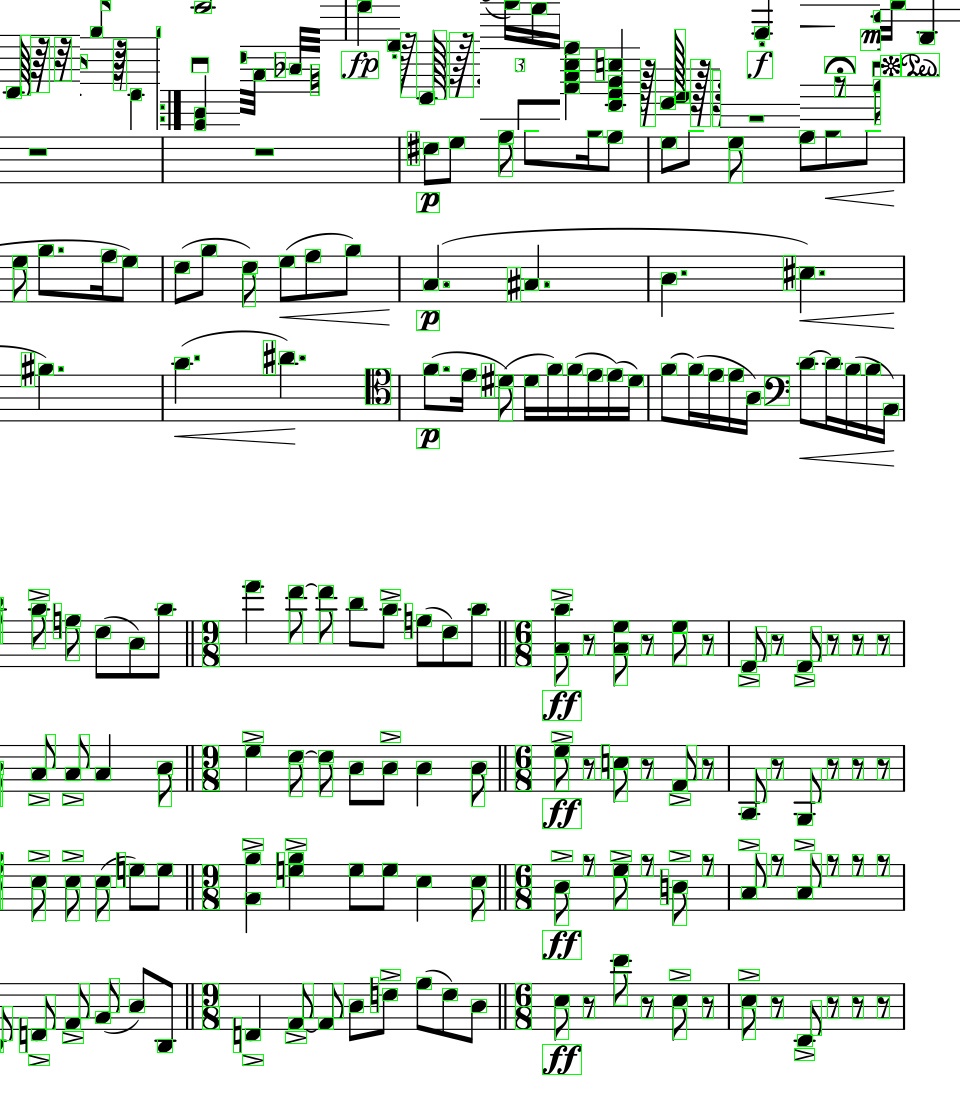}
    \caption{A musical score where $12$ small images have been augmented at the top of $7$ regular staves. The bounding boxes are marked in green.}
    \label{fig:inp_aug}
    \end{center}
    \vspace{-0.8cm}
\end{figure}

Our current answer to this problem is oversampling rare classes by data synthesis, where we locate rare symbols in the dataset, and during training, we append these symbols at the top of the musical sheets (see Figure \ref{fig:inp_aug}). More specifically, we augment each input page in a mini-batch with with $12$ randomly selected synthesized crops of rare symbols (of size $130 \times 80$ pixels) by putting them in the margins at the top of the page. Directions on the choice of the creation of augmented symbols are given on \cite{thilo}. This way, the neural network (on expectation) does not need to wait for more than $10$ iterations to see every class which is present in the dataset. At the same time, we have been experimenting with pre-training the net with fully synthetic scores where the classes are fully balanced and then retraining it on the full \emph{DeepScores} dataset. The two approaches are complementary and preliminary results show improvement, though more investigation is needed: overfitting on extremely rare symbols is still likely, and questions remain regarding how to integrate the concept of patches (in the margins) with the idea of a full page classifier that considers all context.

\subsection{Cached bounding boxes}
The biggest problem of the Deep Watershed Detector (DWD) on a fundamental level is that the bounding box regression is inaccurate. This is possibly due to the fact that convolutional networks produce smooth outputs, but the bounding box map can be very non-smooth. This "smoothing-bias" creates an averaging over all bounding boxes and leads to an overestimation of small bounding boxes and an underestimating of large ones. We currently address this issue by using cached bounding boxes per class as a prediction, being quite accurate for most classes but completely unusable for others. This is a not a satisfactory solution and has to be improved. We are considering multiple approaches including different bounding box encodings in the output layer or usage of the DWD localization as an object proposal system in an R-CNN style detection scheme.

\addtolength{\textheight}{-12cm}   

\end{document}